# A Novel Generative Multi-Task Representation Learning Approach for Predicting Postoperative Complications in Cardiac Surgery Patients

Junbo Shen, B.S.[1,4], Bing Xue, Ph.D.[1,4], Thomas Kannampallil, Ph.D.[1,2,3,4], Chenyang Lu, Ph.D.[1,2,3,4], Joanna Abraham, Ph.D., FACMI, FAMIA[2,3,4]

[1]Department of Computer Science & Engineering, [2]Department of Anesthesiology, [3]Institute for Informatics, Data Science, and Biostatistics, [4]AI for Health Institute

Washington University in St. Louis, St. Louis, MO

**Corresponding author**

Joanna Abraham, PhD, FACMI, FAMIA

Washington University School of Medicine

660 South Euclid Street, Campus Box 8054

St. Louis, MO 63110

joannaa@wustl.edu

314-362-5129

**ABSTRACT**

*Objective*

Early detection of surgical complications allows for timely therapy and proactive risk mitigation. Machine learning (ML) can be leveraged to identify and predict patient risks for postoperative complications. We developed and validated the effectiveness of predicting postoperative complications using a novel surgical Variational Autoencoder (surgVAE) that uncovers intrinsic patterns via cross-task and cross-cohort presentation learning.

*Materials and Methods*

This retrospective cohort study used data from the electronic health records of adult surgical patients over four years (2018 - 2021). Six key postoperative complications for cardiac surgery were assessed: acute kidney injury, atrial fibrillation, cardiac arrest, deep vein thrombosis or pulmonary embolism, blood transfusion, and other intraoperative cardiac events. We compared surgVAE's prediction performance against widely-used ML models and advanced representation learning and generative models under 5-fold cross-validation.

*Results*

89,246 surgeries (49% male, median (IQR) age: 57 (45-69)) were included, with 6,502 in the targeted cardiac surgery cohort (61% male, median (IQR) age: 60 (53-70)). surgVAE demonstrated superior performance over existing ML solutions across all postoperative complications of cardiac surgery patients, achieving macro-averaged AUPRC of 0.409 and macro-averaged AUROC of 0.831, which were 3.4% and 3.7% higher, respectively, than the best alternative method (by AUPRC scores). Model interpretation using Integrated Gradients highlighted key risk factors based on preoperative variable importance.

*Discussion and Conclusion*

Our advanced representation learning framework surgVAE showed excellent discriminatory performance for predicting postoperative complications and addressing the challenges of data complexity, small cohort sizes, and low-frequency positive events. surgVAE enables data-driven predictions of patient risks and prognosis while enhancing the interpretability of patient risk profiles.

**INTRODUCTION**

Over 900,000 cardiac surgical procedures are performed annually in the United States [1,2]. Given the complexity of cardiac surgeries, patients are often at risk for intraoperative and postoperative complications and adverse outcomes, including irregular heartbeats (arrhythmia), excessive blood loss and clots, stroke, acute kidney injury (AKI), delirium, prolonged ventilation, renal failure, and new-onset atrial fibrillation [3–6]. In particular, increased risks of AKI and delirium during and after cardiac surgery have been associated with poor long-term outcomes including reduced cognitive and functional recovery and increased morbidity and mortality [7–11]. Perioperative mortality, depending on the type of cardiac surgery, ranges from 1.8–7.8% [12].

Early detection of these risks can lead to proactive management of clinical signs of deterioration, such as fluid resuscitation, pharmacologic therapies, mechanical ventilation management, and others, potentially preventing adverse outcomes [13]. Although heuristic models for assessing risks exist (e.g., European Cardiac Surgery Evaluation System (EuroSCORE II) [14] or the American Society of Thoracic Surgeons Cardiac Surgery Risk Model (STS score) [15–18]), these are often not personalized or patient-specific.

Machine learning models are able to predict outcomes using large and often complex electronic health record (EHR)-based datasets. For example, models have been developed to predict similar postoperative complications in various surgical cohorts (e.g., [19,20]). However, limited cohort sizes for specialized, infrequent surgeries, which often involve high-dimensional and complex data, have limited the ability to develop robust prediction models [21–24], particularly in small sample clinical contexts such as cardiac surgeries [25–27]. These challenges are further compounded by low outcome event rates, including mortality [27,28], AKI [23,24], cardiac arrest, and deep vein thrombosis/pulmonary embolism. Such limited cohort size coupled with rare events can further lead to brittleness of ML algorithms (e.g., due to model overfitting) [29]. As such, currently developed ML models have failed to elucidate true relationships between preoperative variables and postoperative outcomes. Current studies have selected small subsets of perioperative variables from EHRs (i.e., 25 to 72) to mitigate the challenges of finding complex relationships among preoperative variables [23–25,27,28]. However, such pre-selected methods have failed to incorporate all the rich information needed to accurately characterize patients and their risks.

To address these challenges, we leverage recent advances in generative modeling and develop a novel *surgical Variational Autoencoder* (*surgVAE*) for postoperative cardiac surgery complications. Our approach built upon the earlier explorations of representation learning for Electronic Health Records (EHRs) [29–31].

In this study, our objectives were to: (a) develop and internally validate a novel VAE-based[32] generative model, *surgical Variational Autoencoder* (*surgVAE*), that simultaneously identifies and predicts various postoperative complications of cardiac surgery patients including acute kidney injury (AKI), atrial fibrillation (AF), cardiac Arrest (Arrest), deep vein thrombosis or pulmonary embolism (DVT/PE), blood transfusion, and any other in-hospital cardiac conditions arising postoperatively (Intraop Cardiac Events), and (b) enhance the explainability of postoperative complication predictions based on ranking of the importance of preoperative input variables.

**METHOD**

*Setting and Data Sources*

Data were obtained from the EHR for all adult surgical patients undergoing surgical procedures at Washington University School of Medicine and Barnes-Jewish Hospital over four years (2018—2021). This academic medical center serves a diverse population in Missouri and Illinois and uses the Epic Electronic Health Record (Epic Systems, Verona, WI). Approval from the institutional review board of Washington University School of Medicine with a waiver of patient consent was obtained [33]. Transparent Reporting of a Multivariable Prediction Model for Individual Prognosis or Diagnosis (TRIPOD-AI) guidelines were used for reporting this study.

*Data*

Preoperative data and target patient outcomes related to postoperative complications were retrieved from EHR. Preoperative variables included patient demographics, medical history and acuity, physiological measurements, laboratory measurements, and surgical notes describing the planned procedure (*see* Supplementary Table 1 for complete list). Retrieved postoperative complications included AKI status, atrial fibrillation (AF), cardiac arrest (arrest), deep vein thrombosis or pulmonary embolism (DVT/PE), blood transfusion, and any other in-hospital cardiac conditions arising postoperatively (intraop cardiac events).

*Data Processing*

There were 163 preoperative variables. Missing rates of variables ranged between 0 to 97.4%, with an average missing rate of 30.6% before imputation. Each preoperative variable was normalized, and missing data were imputed as zero. The planned surgery description consisted of 3204 unique words and each surgery has 8.4 words on average (with a standard deviation of 6.9 words); it was tokenized and transformed into textual embedding with 500 numbers using a continuous-bag-of-words (CBOW) model [34]. In total, there were 663 input features for ML models.

All postoperative complications (outcomes) were categorized as either positive or negative. Positive outcomes indicated an elevated postoperative risk while negative outcomes signified the absence of the corresponding risk. Blood transfusion, measured as volume in mL, was categorized into positive and negative classes: any postoperative blood transfusion of greater than 0 mL was classified as positive, the absence of blood transfusions was classified as negative. AKI status was recorded as four classes (0, 1, 2, 3) and categorized into positive (1, 2, 3) and negative (0). AF, arrest, DVT/PE, and intraop cardiac events were originally recorded as two classes (positive and negative).

AKI was determined from a combination of lab values (serum creatinine) and dialysis event records, using KDIGO criteria with 3 stages. Structured anesthesia assessments, laboratory data, and billing data indicating baseline end-stage renal disease (ESRD) were used as exclusion criteria for AKI. Diagnosis codes for AF, arrest, DVT, PE, and intraop cardiac events are provided in Supplementary Tables 2- 6. Diagnosis codes for outcome definitions are not billing codes created at the end of a hospitalization, but rather time-stamped problem entries during the hospitalization for addressing acute problems.

*Outcomes*

Target outcomes included the following postoperative complications: AKI status, AF, arrest, DVT/PE, blood transfusion, and any other cardiac conditions arising postoperatively (intraop cardiac events). These complications were selected as they are potentially modifiable during the postoperative period, and early identification could mitigate their negative impact on outcomes [3,7,23–27].

*Baseline Models*

We applied widely used ML models in clinical predictions, including eXtreme Gradient Boosting (XGBoost), Deep Neural Network (DNN), Multi-Task Deep Neural Network (Multi-task DNN), Logistic Regression (LR), Random Forest (RF) and Gradient Boosting Machine (GBM). These models have been used to predict cardiac surgery-associated complications with high validity and accuracy[23,24,27]; XGBoost and RF models had the highest accuracy in most perioperative complications [23,24,27,35]. We implemented XGBoost using the XGBoost package, Multi-task DNN using Pytorch, and other ML models in Scikit-learn package in the Python environment [36–39].

In addition to above models, we also extended our approach to the recent state-of-the-art representation learning and meta-learning algorithms, which have shown to be promising solutions in other prediction tasks. These models included Variational Autoencoder (VAE), Beta Total Correlation Variational Autoencoder (Beta TC VAE), Factor Variational Autoencoder (FactorVAE), clinical Variational Autoencoder (cVAE), Prototypical Network, and Model-Agnostic Meta-Learning (MAML), with major differences in their respective objective functions and modeling strategies [29,32,40–43] (*see* Supplementary Note 3 for comparisons). VAE, Beta TC VAE, FactorVAE, and cVAE used shared representations for downstream outcome predictions, and prediction heads were used to draw predictions from the latent encoding. Prototypical Network and MAML applied adapting/finetuning to the entire pre-trained model for each specific outcome prediction. Although most of these advanced generative models have not been used in postoperative predictions, a recent study has shown that generative models could achieve state-of-the-art performance in predicting postoperative outcomes such as delirium and length of stay in operating room [29]. Implementation details are shown in Supplementary Table 7. *See* Supplementary Note 1 for details of training parameters.

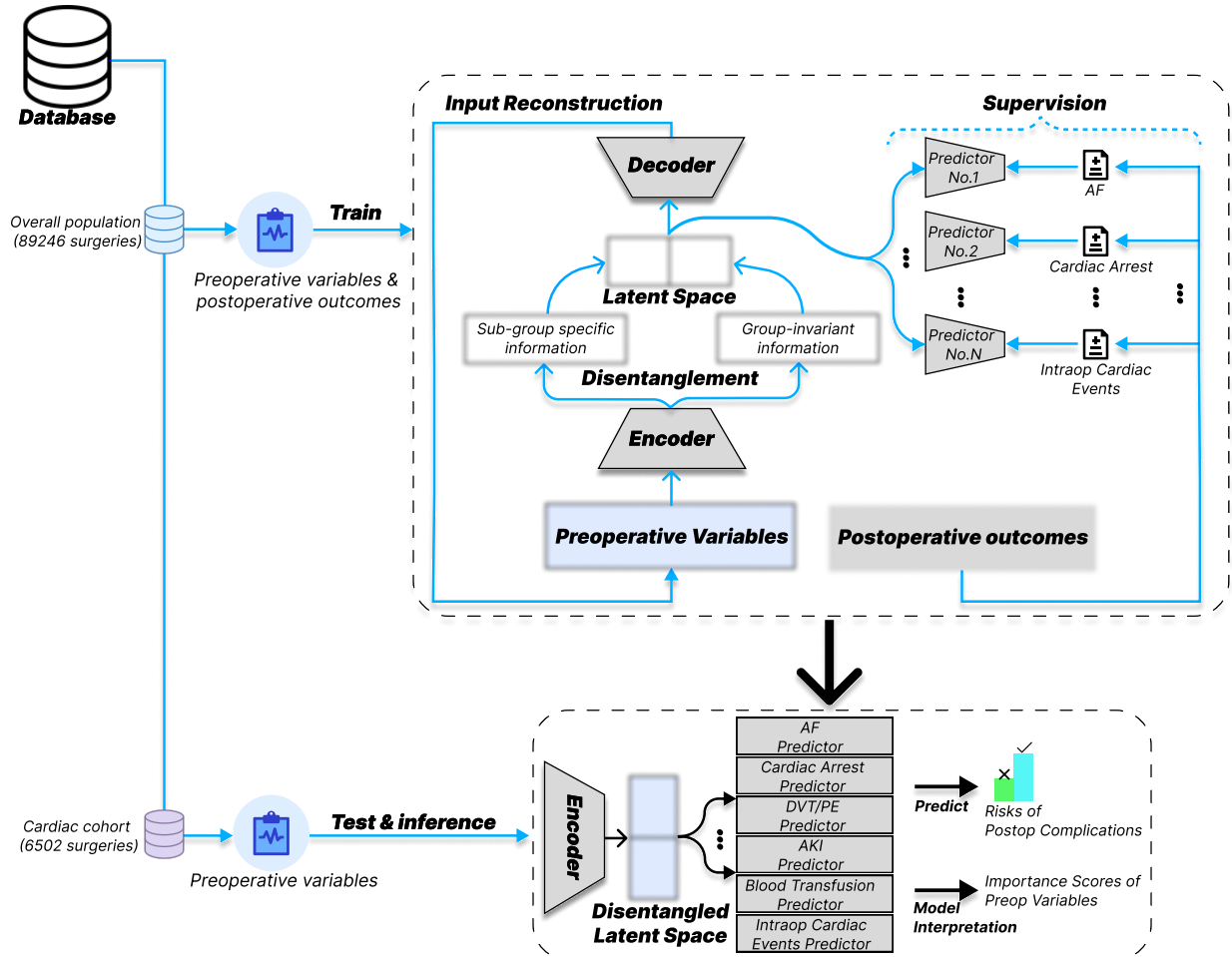

**Figure 1.** Illustration of surgVAE. surgVAE is a unified model with auxiliary predictors tailored for N = 6 complications (AF, arrest, DVT/PE, AKI, blood transfusion, intraop cardiac events), enabling simultaneous prediction across different complications. Latent space is a low-dimensional representation of patient characteristics, where each dimension represents a single aspect of characteristics. It is disentangled into sub-group specific and group-invariant latent space for all surgeries to optimally represent characteristics derived from postoperative variables from different surgery cohorts. Target complications are used during training to supervise the learning process.

*Proposed surgVAE*

As depicted in **Figure 1**, surgVAE integrated several new salient features compared to other state-of-the-art representation learning models. Unlike standard VAE models, it included supervised learning tasks (via additional prediction heads) with self-supervised learning tasks such that the lower-dimensional representation of high-dimensional inputs was guided by supervised tasks (regularization and reconstruction) as well as prediction tasks (for postoperative complications). With a multi-task architecture, the learned knowledge in predicting one complication was shared across the latent space and prediction heads for better prediction of other tasks. In order to better discover the intrinsic information between inputs and outputs, surgVAE learned from non-cardiac cohorts through a careful design of its latent space, which was divided into a group-specific subspace to represent unique patient characteristics of a certain cohort (e.g., cardiac surgeries) and a group-invariant subspace representing mutual information among different surgery cohorts. The introduction of the group-invariant space preserved shareable knowledge to mitigate the limitations of sample sizes and number of positive events among cardiac surgeries. We enabled disentanglement between two subspaces by maximizing the statistical differences in the group-specific subspace and minimizing the statistical differences in the group-invariant space.

*Architecture of surgVAE*

surgVAE is constituted of an encoder $q_\phi(z|x)$, a decoder $p_\theta(x|z)$. The encoder transforms the 663-dimensional input $x$ to a latent representation $z$ by first encoding the sequential variables into 2-dimensional tokens using a feature encoder, followed by two attention blocks and fully connected layers. The decoder reconstructs the synthetic $\hat{x}$ from the latent representation $z$ using three fully connected layers. Additionally, surgVAE includes six auxiliary classifiers, each designed to predict a specific postoperative complication, that take the direct output from the encoder as input and pass them through three fully connected layers for predictions (*see* Supplementary Figures 1, 2 for illustrations).

To train surgVAE, we extend the Evidence Lower Bound (ELBO) from the vanilla VAE by incorporating additional losses. The ELBO objective in vanilla VAE[32] has two components: a reconstruction term $\mathcal{L}_{\text{recons}}$ and a regularization term $\mathcal{L}_{KLD}$, defined as:

$$\mathcal{L}_{ELBO}(x) = \mathcal{L}_{\text{recons}}(x) + \mathcal{L}_{KLD}(x) \quad (1)$$
$$\equiv \mathbb{E}_{q_\phi(z|x)}[\log p_\theta(x|z)] - \text{KL}(q_\phi(z|x)\|p(z))$$

Here, KL denotes the Kullback-Leibler divergence.

We introduce Total Correlation (TC) Loss, inspired by Beta TC VAE[40], to encourage disentanglement of the latent space into two components: $z_1$, representing group-invariant space across all surgeries, and $z_2$, representing sub-group specific latent space:

$$\mathcal{L}_{\text{TC}} = -\text{KL}(q_\phi(z) \| \prod_{j=1} q_\phi(z_j)) \quad (2)$$

To minimize discrepancies in the latent space $z_1$ across different surgery groups, thereby improving the representation of shared knowledge, we apply Distribution Matching Loss using Maximum Mean Discrepancy (MMD) introduced in infoVAE[44]. Letting $k(\cdot,\cdot)$ be the Gaussian kernel, the distribution matching loss is:

$$\mathcal{L}_{MMD} = \mathbb{E}_{p(z_1),p(z_1')}[k(z_1,z_1')] + \mathbb{E}_{q(z_1),q(z_1')}[k(z_1,z_1')] - 2\mathbb{E}_{p(z_1),q(z_1')}[k(z_1,z_1')] \quad (3)$$

Additionally, we introduce Contrastive Loss, which helps discriminate between different surgery groups by minimizing statistical differences of latent spaces $z_2$ from the same surgery group and maximizing those from different surgery groups. Letting positive pairs $\mathcal{P}$ be any pairs of latent space $z_2$ from the same surgery group inside the batch of size $N$, and negative pairs $\mathcal{N}$ be any pairs of latent space $z_2$ from different surgery groups inside the batch of size $N$, the contrastive loss is:

$$\mathcal{L}_{\text{contrastive}} = \frac{1}{N} \sum_{(i,j) \in \mathcal{P}} \|z_2^i - z_2^j\|^2 + \frac{1}{N} \sum_{(i,j) \in \mathcal{N}} \max\left(0, \text{margin} - \|z_2^i - z_2^j\|\right)^2 \qquad (4)$$

To better preserve predictive information in the latent space, we apply cross-entropy loss between $C = 6$ number of target complications $y$ and auxiliary classifiers' predictions $\hat{y}$:

$$\mathcal{L}_{\text{prediction}} = \frac{1}{C} \sum_{i=1}^{C} \text{CrossEntropy}(y_i, \hat{y}_i) \qquad (5)$$

The overall objective of **eqns (1)- (5)** is to ensure that surgVAE learns a disentangled and interpretable latent space, aligns representations across surgery groups, and remains effective for multi-outcome predictions (*see* Supplementary Note 1 for details).

*Model Training and Evaluation*

For an unbiased assessment of all models, we stratified target cardiac surgery cohort into 5 folds, and reported the average cross-validated metrics in all subsequent experiments. Stratification was based on the class distribution of the most imbalanced outcome (cardiac arrest with positive rate < 1%) while also ensuring that other outcomes were nearly evenly distributed across the folds. During cross-validation, one fold served as the test set to calculate the model performance, and the remaining four folds constituted the training set.

All methods were trained and tested under the cross-validation settings, and all VAE models (Vanilla VAE, Factor VAE, Beta TC VAE, and our surgVAE) shared the same encoder-decoder architecture design. Due to the rareness of positive events in the cardiac cohort, non-cardiac surgery cases were added to the training stage of all baseline models as well as proposed surgVAE for data augmentation. This helped VAE models with a more generalizable latent representation of patient characteristics during self-supervised learning and helped Meta-Learning Models (MAML and Prototypical Network) to learn the transferable information from a broader surgical population to cardiac surgery patients. In all cases, models were evaluated exclusively on the cardiac surgery data in the test fold.

Two key performance measures, the area under the precision-recall curve (AUPRC) and the area under the receiver operating characteristic curve (AUROC), were recorded in each iteration as they report the overall model performance regardless of threshold setting. The average precision-recall curves (PRC) and receiver operating characteristic (ROC) curves were plotted for the best-performing models.

*Model Interpretability*

To identify the contribution of each input preoperative variable to final predictions, we applied the Integrated Gradient [45] method to derive complication-specific importance scores for each preoperative variable. This gradient-based feature attribution method is widely used to interpret complex deep-learning models as it calculates importance scores based on the gradient of the output compared to input features. For example, if the importance score for a preoperative variable is high, then slight deviations in that variable could cause large deviations in output results.

*Clustering*

To visually compare the surgVAE's latent space with the original input space (preoperative variables), we used t-distributed stochastic neighbor embedding (t-SNE) [46]. Each point in the t-SNE plot represented a cardiac surgery case, colored by postoperative outcomes (e.g., AF, AKI, blood transfusion). The distribution difference in the t-SNE plots illustrated how surgVAE aggregated similar cases and clustered them in the latent space compared with the raw variables, where better separation based on shared risk factors and outcomes revealed more meaningful clusters that could potentially improve outcome predictions.

**RESULTS**

*Study Population*

A total of 89,246 surgeries (median [InterQuartile Range, IQR] age, 57 [45, 69] years; 49% male) were included, with 6502 cardiac surgeries (age, 60 [53, 70] years; 61% male). **Table 1** shows the characteristics of patients from the overall population and cardiac cohort in the study. The positive event rates for AF, arrest, DVT/PE, AKI, blood transfusion, and intraop cardiac events were 25.87%, 0.40%, 2.17%, 32.16%, 31.04%, and 4.51%, respectively (*see* Supplementary Table 8 for details).

**Table 1.** Characteristics of the overall cohort and cardiac cohort. Categorical variables represented as frequency (%). Continuous variables represented as median (25th percentile, 75th percentile).

| Features | | Overall Population (n= 89246) | Cardiac Cohort (n=6502) |
|---|---|---|---|
| Age (years) | | 57 (45, 69) | 60 (53, 70) |
| Male sex | | 44078 (49) | 3961 (61) |
| White race | | 66140 (74) | 5521 (85) |
| Height (cm) | | 170 (163, 178) | 172 (165, 180) |
| Weight (kg) | | 86 (69, 100) | 84 (68, 97) |
| ASA physical status | 1 | 3633 (4.1) | 17 (0.3) |
| | 2 | 31944 (36) | 574 (8.8) |
| | 3 | 38423 (43) | 2361 (36) |
| | 4 | 8424 (9.4) | 2788 (43) |
| | >4 | 384 (0.4) | 121 (1.9) |
| | Missing | 6438 (7.2) | 641 (9.9) |
| ASA emergency status | | 3838 (4.3) | 341 (5.2) |
| Hypertension | | 48704 (55) | 4174 (64) |

| | | |
|---|---|---|
| Coronary artery disease | 13865 (16) | 2381 (37) |
| Prior myocardial infarction | 8782 (9.8) | 1450 (22) |
| Congestive heart failure | 9332 (10) | 2114 (33) |
| Atrial fibrillation | 10304 (12) | 2021 (31) |
| Pacemaker | 3360 (2.3) | 829 (13) |
| Prior stroke or transient ischemic attack | 8155 (9.1) | 844 (13) |
| Peripheral artery disease | 10699 (12) | 1303 (20) |
| Deep venous thrombosis or pulmonary embolism | 9106 (10) | 866 (13) |
| Diabetes | 22346 (35) | 2344 (36) |
| Outpatient insulin use | 5991 (6.7) | 448 (6.9) |
| Chronic kidney disease | 12999 (15) | 1471 (23) |
| Ongoing dialysis | 4648 (5.2) | 464 (7.1) |
| Pulmonary hypertension | 5519 (6.2) | 957 (15) |
| Chronic obstructive pulmonary disease | 10157 (11) | 1445 (22) |
| Asthma | 10346 (12) | 627 (9.6) |
| Obstructive sleep apnea | 15802 (18) | 1312 (20) |
| Cirrhosis | 4076 (4.6) | 368 (5.7) |
| Any cancer | 30491 (34) | 2294 (35) |
| Gastro-esophageal reflux | 32888 (37) | 2718 (42) |
| Anemia | 21010 (24) | 1626 (25) |
| Coombs-positive | 2118 (4.6) | 193 (3.0) |
| Dementia | 2745 (3.1) | 117 (1.8) |
| Ever-smoker | 32061 (70) | 3070 (47) |

*Prediction of Target Postoperative Complications after Cardiac Surgery*

The surgVAE model showed superior performance in predicting postoperative complications after cardiac surgery. AUROCs [Mean (Standard Error)] of surgVAE were: AF [0.838 (0.013)], arrest [0.773 (0.056)], DVT/PE [0.831 (0.021)], AKI [0.864 (0.006)], blood transfusion [0.874(0.008)], and intra cardiac events [0.807 (0.031)]. Corresponding AUPRCs [Mean (Standard Error)] were: AF [0.633 (0.031)], arrest [0.018 (0.013)], DVT/PE [0.134 (0.057)], AKI

[0.761 (0.014)], blood transfusion [0.737 (0.012)], and intra cardiac events [0.170 (0.040)]. The macro-averaged AUPRC and macro-averaged AUROC over all six outcomes were 0.409 and 0.831, respectively.

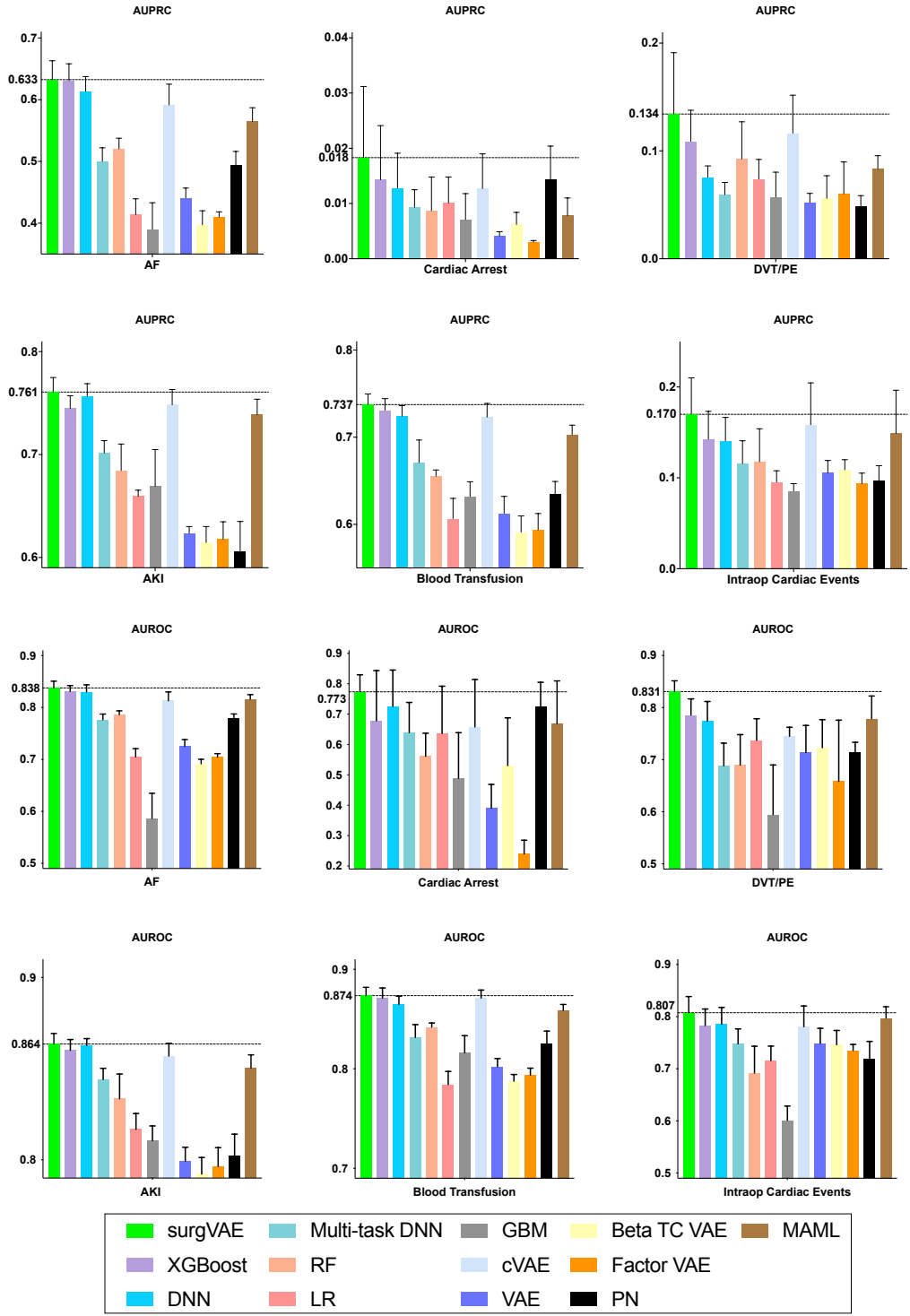

**Figure 2.** Cross-validated prediction performance of surgVAE and competitive methods measured by AUPRC and AUROC. Error bars represent standard deviations based on a sample size of 5. Evaluated outcomes include AF: atrial fibrillation; arrest; DVT/PE: deep vein thrombosis or pulmonary embolism; AKI: post-acute kidney injury status; blood transfusion; intraop cardiac events: new cardiac conditions arising postoperatively. Methods include XGBoost: eXtreme Gradient Boosting; DNN: multi-layer deep neural network; Multi-task DNN: multi-task multi-layer deep neural network; LR: logistic regression; RF: random forest; GBM: gradient boosting machine; VAE: variational autoencoder; Beta TC VAE: beta total correlation variational autoencoder; Factor VAE: factor variational autoencoder; cVAE: clinical Variational Autoencoder; PN: prototypical network; MAML: model-agnostic meta-learning.

*Validation of Prediction Model*

surgVAE outperformed all other candidate models in terms of both AUROC and AUPRC for all six postoperative prediction tasks, as shown in **Figure 2** (*see* Supplementary Table 9 for details). The next best-performing algorithm after surgVAE (ranked by macro-averaged AUPRC) was XGBoost (macro-averaged AUPRC = 0.395; macro-averaged AUROC = 0.801). surgVAE achieved higher macro-averaged AUPRC by 3.4% and macro-averaged AUROC by 3.7%. The next best-performing representation learning algorithm after surgVAE was cVAE (macro-averaged AUPRC = 0.392; macro-averaged AUROC = 0.787). The next best-performing VAE model after surgVAE was cVAE, and surgVAE achieved a 4.3% increase in macro-averaged AUPRC and 5.6% in macro-averaged AUROC. **Figure 3a** shows the receiver operating characteristic (ROC) curves and precision-recall curves for surgVAE across all six postoperative complications. The ROC curves were consistently strong for all outcomes, highlighting surgVAE's robustness.

We further reported the cross-validated specificity, precision, and accuracy for each complication by fixing the sensitivity at 0.85. For comparison, we also reported the metrics for XGBoost (best alternative model) at the same sensitivity. (*See* Supplementary Table 10 for complete results). From the comparison, surgVAE achieved higher performance than XGBoost in each metric: AF accuracy = 0.696, precision = 0.447, and specificity = 0.644; AKI accuracy = 0.759, precision = 0.594, and specificity 0.714; blood transfusion accuracy = 0.769, precision =

0.589, and specificity =0.732; intraop cardiac events accuracy = 0.620, precision = 0.105, and specificity = 0.607; DVT/PE accuracy = 0.655, precision = 0.051, and specificity = 0.650; arrest accuracy = 0.530, precision = 0.008, and specificity = 0.528.

Moreover, we generated the t-SNE plots of the original input space (preoperative variables) and the latent space generated by surgVAE. AF, AKI, and blood transfusion were used for visualization because these outcomes have more balanced class distributions (higher positive rate), providing sufficient data points to visualize. The latent space of surgVAE revealed distinct clusters, referred to as "phenotyping effects"[29] (**Figure 3b**), which strongly correlated with postoperative outcomes. This demonstrates surgVAE's capability to discern meaningful patterns in complex preoperative variables, which is crucial for predicting postoperative risk heterogeneity (*see* Supplementary Note 2 for details).

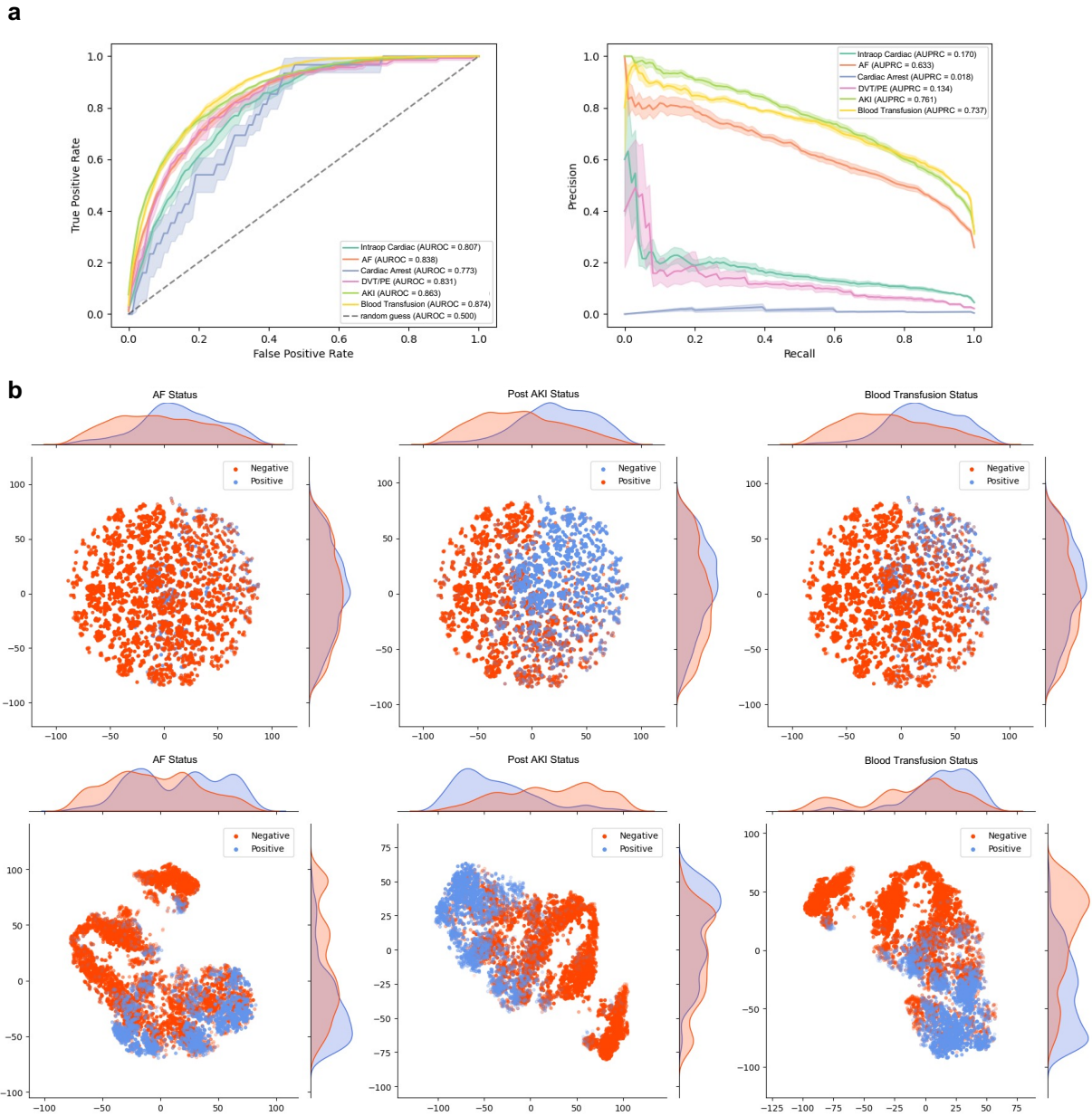

**Figure 3.** Performance and clustering effects in cardiac surgery outcome predictions. **a**, The best-performing model (surgVAE) in postoperative outcome predictions. Left: right: receiver operating characteristic (ROC) curves; Right: precision-recall curves (PRC) The solid line represented the average and the shades represented their respective SEs at each point during cross-validation. **b**, t-SNE plots of cardiac patients, colored by postoperative outcomes. Top: inseparable clusters from the original preoperative variables; bottom: well-separated clusters in representation (latent) space of surgVAE. Each scatter point represents a cardiac surgery case, colored by its corresponding postoperative outcomes.

*Model Interpretation*

Our analysis revealed that the prediction model showed consistent discriminant validity for each of the complications. Postoperative complication-specific interpretation of surgVAE is presented in **Figure 4**. As most preoperative variables have relatively small importance scores, we selected the top 10 preoperative variables with the highest importance scores for each complication.

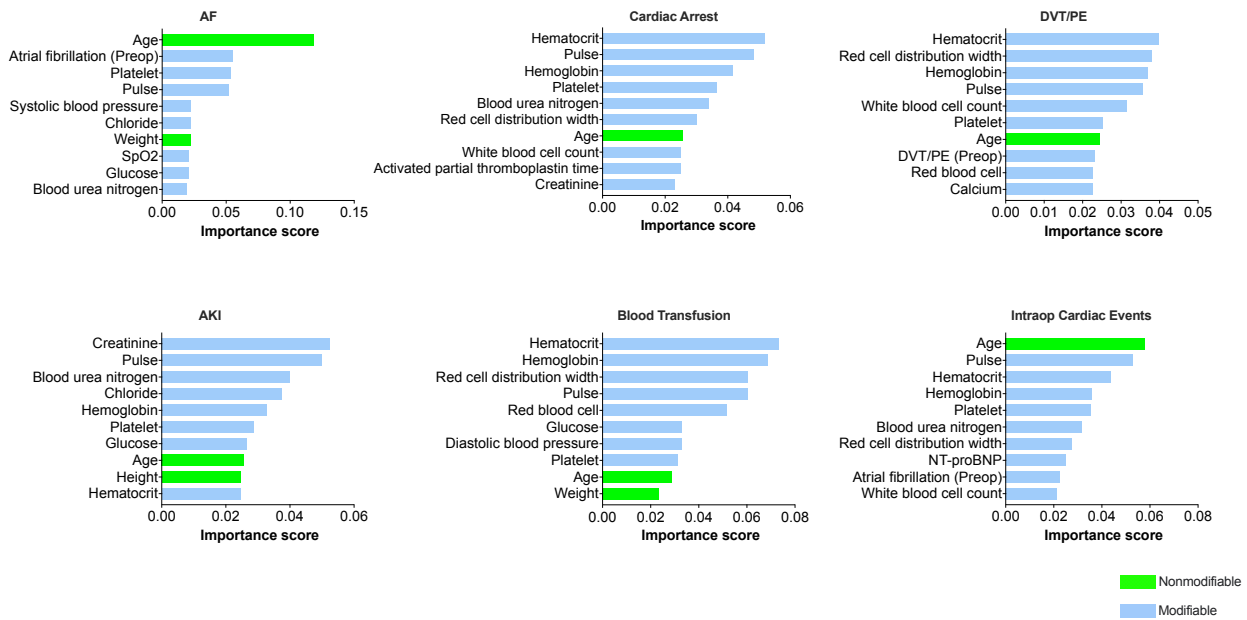

**Figure 4.** Complication-specific interpretation for surgVAE. The top 10 features with the highest importance scores are selected for each postoperative complication, including both modifiable and nonmodifiable features. Importance scores are calculated using the integrated gradient and normalized to present the percentage of contributions of each preoperative variable for accurate predictions on target postoperative complications.

Among non-modifiable features, "age" stood out as a significant factor across many categories and served as a baseline indicator rather than a target for intervention. It was the most critical predictor for AF, with an importance score of 11.9%, and remained significant in other outcomes like arrest and DVT/PE. This reflected the universal impact of patient age on postoperative outcomes. For modifiable features, preoperative variables related to blood composition, such as hematocrit, hemoglobin, and platelet count, appeared as key predictors in several categories, which indicated the critical role of patients' hematologic status in the risk of postoperative complications. "Pulse" was a top modifiable feature for many complications, including AF, arrest, DVT/PE, and AKI, which might be indicative of its role as a general marker of hemodynamic stability and cardiovascular risk. For DVT/PE, "Red cell distribution width" stood out, suggesting the importance of variations in red cell size for this complication. Meanwhile, "creatinine" levels were important for AKI, which was expected as it directly correlated with kidney function. Vital signs and laboratory values are particularly interesting cases for understanding modifiability because even though they can be changed, they also reflect potentially un-recorded severity of other medical problems. For example, "hematocrit" can be modified by preoperative iron therapy or transfusion. "pulse" can be decreased by beta-blocking medications and a variety of other medications, or if dangerously low increased by chronotropic agents such as dopamine and glycopyrrolate. Detailed interpretation results are provided in Supplementary Table 11.

**DISCUSSION**

We developed and validated a novel surgical Variational Autoencoder (surgVAE) to identify cardiac surgery patients who are at risk for postoperative complications. surgVAE outperformed widely used machine learning models in clinical predictions, as well as advanced meta-learning approaches (Prototypical Networks and MAML) and generative models (VAE, Beta TC VAE, Factor VAE, and cVAE). Compared with these methods, surgVAE introduced both supervised learning strategies and self-supervised learning strategies with disentangled latent subspaces, enabling the model to extract the key patterns between massive preoperative variables and postoperative complications, meanwhile discovering more shareable knowledge by utilizing the non-cardiac surgery cases and different prediction tasks.

*Strengths of the Proposed surgVAE Risk Model*

Our extensive experiments showed surgVAE with consistently better prediction performance than existing methods. This highlights the promise of leveraging advanced generative modeling techniques in perioperative cardiac surgery care that can simultaneously identify various postoperative risk factors. Through knowledge sharing, the risk prediction of each postoperative complication benefited from shared representation in the latent space, where this latent space is co-supervised by multiple key outcomes across various surgery cohorts. Such design enabled surgVAE's versatility in providing a holistic assessment of postoperative risks and could potentially extend to more complications and other cohorts-of-interest.

surgVAE incorporates all available variables—even those with high missingness rates—through latent representations that enhance predictive accuracy across multiple complications (*see* Supplementary Table 12 for missingness rates in preoperative variables). By leveraging a single model for multiple outcomes, surgVAE represents a foundational model approach that reduces the computational overhead of building multiple bespoke models for various tasks. This brings surgVAE unique advantages over widely used machine learning models such as XGBoost, which is a bespoke model trained on a particular dataset for a specific task.

surgVAE can utilize the information in other cohorts (non-cardiac surgeries) to address the sparsity of positive events, making it particularly favorable for clinical problems facing limited data size or when data collection is expensive. In fact, such "few-shot learning" capabilities have been proven in various generative models, such as VAE, transformer, and GPT in image generation and large language models, and are extensively used in ChatGPT [47]. In our study, we validated the few-shot learning by utilizing the non-cardiac cases and their label information to promote knowledge distillation and sharing, and used the learned information in the cardiac surgery cases that had different patient characteristics and label distributions.

*Interpretability of surgVAE Risk Model*

To further enhance the clinical utility of our model, we integrated interpretation for surgVAE. We incorporated integrated gradients [45] for model interpretation, particularly suited for complex deep learning models. We offered a scalable and practical approach over traditional methods like SHAP [48], which may struggle with the computational intensity and inefficiency of calculating exact Shapley values for complex models with high-dimensional inputs. Our interpretability results underscored that certain preoperative variables are more critical than others in predicting specific postoperative outcomes, which aligned with established clinical observations. Careful monitoring of these identified variables could also be beneficial in clinical decision-making for correct intervention to mitigate risks of postoperative complications.

*Clustering effects of surgVAE*

We created t-SNE plots to project surgVAE's learned latent space to visually interpret the relationships between cardiac surgery cases and their corresponding postoperative outcomes. Our plots highlighted the clustering effects with respect to postoperative complications where cardiac surgeries with the same postoperative outcome were grouped in the same cluster. The distinct clusters in surgVAE's latent space demonstrated the model's capability to learn complex relationships between high-dimensional preoperative variables and outcomes through latent encoding, which otherwise would not be visualizable from the original input space.

*Practical Application of the surgVAE Risk Model*

*surgVAE's* superior prediction performance in all target complications demonstrated its promising application in perioperative decision-making. To highlight the clinical utility and translational impact of surgVAE in perioperative care, we presented a case example of a patient with positive postoperative complications for atrial fibrillation, blood transfusion, and new cardiac conditions (intraop cardiac events).

> <u>Example scenario</u>: A 77-year-old patient with a history of anemia, asthma, coronary artery disease, chronic kidney disease, deep vein thrombosis, gastro-esophageal reflux disease, hypertension, myocardial infarction, obstructive sleep apnea, pulmonary hypertension, and diabetes was admitted for cardiac surgery. The patient's weight was 87.6 kg and height was 63 inches, indicating a BMI that suggests obesity, a common risk factor for many conditions listed. The patient had elevated creatinine levels at 1.97 mg/dL, a hemoglobin A1c of 6.5%, a significantly high neutrophil count at 16.1 K/µL, low hemoglobin (9.8 g/dL) and hematocrit (28.9%), a low platelet count of 70 K/µL, an N-terminal pro-b-type natriuretic peptide (NT-proBNP) level of 903 pg/mL.

For this patient, surgVAE accurately predicted the occurrence of positive postoperative complications (AF, blood transfusion, intraop cardiac events), with high confidence scores of 0.844, 0.924, and 0.699, respectively. We provide a complication-specific interpretation. The primary risk factors for AF included the patient's preoperative AF condition, age, and platelet count. For blood transfusions, the critical predictors were hematocrit, hemoglobin, and platelet count. The major factors for intraop cardiac events were NT-proBNP level, platelet count, and red cell distribution width.

*Implications for Perioperative Practice:* Insights from this study can inform the development of ML-augmented clinical decision support system (CDSS) to support clinicians in care management and treatment decisions. Integration of advanced representation learning and generative modeling into current CDSS can have several implications: reduce errors in perioperative care by providing preemptive alerts of preventable postoperative complications to clinicians; support care coordination and tailored communication during intraoperative and postoperative handoffs [49–51] to foster anticipatory guidance, resilience to preventable complications, social interaction (e.g., shared understanding), and information processing [52]. While many of these benefits may be achievable with various ML methods, surgVAE offers distinct advantages in dealing with limited cohort size and making multi-outcome predictions simultaneously. The generative nature enables surgVAE to learn complex patterns from diverse patient populations, allowing it to better generalize to new or unseen patient groups and clinical settings. These strengths enable surgVAE to provide actionable insights even when patient data is incomplete or subject to domain shifts, making it a better tool for enhancing clinical decision-making.

This study serves as a *proof-of-concept* highlighting an innovative methodological framework to enable data-driven predictions of patient risks and prognosis, while enhancing the interpretability of patient risk profiles.

We acknowledge our study limitations. *First*, we only used data from a single academic medical center, which can limit its generalizability. However, this study serves as a proof-of-concept and can offer a foundational understanding of the potential for using advanced ML for surgical risk predictions, particularly with surgVAE leveraging the overall surgical cohort while targeting the high-risk cardiac surgery cohort. *Second*, as a deep learning model, surgVAE requires more sophisticated implementation than traditional machine learning models, however, as deep learning continues to advance and become more accessible, we anticipate that the deployment of models like surgVAE will be increasingly practical in clinical settings. Finally, only preoperative variables were used to predict postoperative complications. Given the complexity of cardiac surgery and various types of perioperative variables collected at other medical centers, future research should extend this research and explore ways to integrate different types and modalities of data from multiple sites to enhance further robustness and generalizability of surgVAE across populations/settings.

## CONCLUSION

This study demonstrates the potential of advanced machine learning to significantly enhance the prediction and prevention of postoperative complications in cardiac surgery, supporting more accurate and interpretable risk assessments, and enabling clinicians to make better-informed decisions. Our methodological innovations are three-fold: (1) adapting machine learning to effectively handle small patient cohorts with low event rates, (2) integrating recent advances in deep learning into a unified multi-task solution for perioperative care, and (3) overcoming limitations of previous approaches, significantly outperforming state-of-the-art models and remaining highly interpretable.

## AUTHOR CONTRIBUTIONS

JS, BX, and JA wrote the manuscript draft. JS and BX processed the EHR data and did the statistical analysis. JS implemented the machine-learning methods. JS, BX, and JA had full access to all the data in the study and take responsibility for the integrity of the data and the accuracy of the data analysis. JS, BX, CL, and JA contributed to the conception of the work and design of the study. JS, BX, and JA contributed to the acquisition, analysis, or interpretation of data. JA and TK provided administrative, technical, or material support. CL and JA supervised the study. All authors contributed to the revision of the manuscript.

## ACKNOWLEDGEMENTS

Members of the TECTONICS Study Group who contributed to data acquisition or other support of the implementation are provided in the Supplement.

## SUPPLEMENTARY INFORMATION

Supplementary Notes, Tables, and Figures have been attached and provided alongside the manuscript submission to the journal.

## FUNDING

The study received no funding.

## DECLARATION OF INTERESTS

## DATA AVAILABILITY

The Washington University Human Research Protection Office did not permit sharing of patient-level data due to enrollment with a waiver of consent. To facilitate a better understanding of the data, we have provided sample synthetic data, which is available in open access figshare (https://figshare.com/s/28baa2cc457dab33a454).

## CODE AVAILABILITY

The code being used in the current study for developing the method is provided via GitHub at https://github.com/ai4biomedicine/surgVAE.

## REFERENCES

1   D'Agostino RS, Jacobs JP, Badhwar V, *et al*. The Society of Thoracic Surgeons Adult Cardiac Surgery Database: 2018 Update on Outcomes and Quality. *Ann Thorac Surg*. 2018;105:15–23. doi: 10.1016/j.athoracsur.2017.10.035

2   Research iData. How Many Cardiac Surgeries Are Performed Each Year? - New Study by iData Research. IData Res. 2023. https://idataresearch.com/over-900000-cardiac-surgeries-performed-every-year-in-the-united-states/ (accessed 12 August 2024)

3   Thiele RH, Isbell JM, Rosner MH. AKI associated with cardiac surgery. *Clin J Am Soc Nephrol CJASN*. 2015;10:500–14. doi: 10.2215/CJN.07830814

4   O'Brien SM, Feng L, He X, *et al.* The Society of Thoracic Surgeons 2018 Adult Cardiac Surgery Risk Models: Part 2-Statistical Methods and Results. *Ann Thorac Surg*. 2018;105:1419–28. doi: 10.1016/j.athoracsur.2018.03.003

5   Cornwell LD, Omer S, Rosengart T, *et al.* Changes over time in risk profiles of patients who undergo coronary artery bypass graft surgery: the Veterans Affairs Surgical Quality Improvement Program (VASQIP). *JAMA Surg*. 2015;150:308–15. doi: 10.1001/jamasurg.2014.1700

6   Mehaffey JH, Hawkins RB, Byler M, *et al.* Cost of individual complications following coronary artery bypass grafting. *J Thorac Cardiovasc Surg*. 2018;155:875-882.e1. doi: 10.1016/j.jtcvs.2017.08.144

7   Wang Y, Bellomo R. Cardiac surgery-associated acute kidney injury: risk factors, pathophysiology and treatment. *Nat Rev Nephrol*. 2017;13:697–711. doi: 10.1038/nrneph.2017.119

8   Sprung J, Abdelmalak B, Gottlieb A, *et al.* Analysis of risk factors for myocardial infarction and cardiac mortality after major vascular surgery. *Anesthesiology*. 2000;93:129–40. doi: 10.1097/00000542-200007000-00023

9   Hobson CE, Yavas S, Segal MS, *et al.* Acute kidney injury is associated with increased long-term mortality after cardiothoracic surgery. *Circulation*. 2009;119:2444–53. doi: 10.1161/CIRCULATIONAHA.108.800011

10  Goldfarb M, Drudi L, Almohammadi M, *et al.* Outcome Reporting in Cardiac Surgery Trials: Systematic Review and Critical Appraisal. *J Am Heart Assoc*. 2015;4:e002204. doi: 10.1161/JAHA.115.002204

11  Gaudino M, Rahouma M, Di Mauro M, *et al.* Early Versus Delayed Stroke After Cardiac Surgery: A Systematic Review and Meta-Analysis. *J Am Heart Assoc*. 2019;8:e012447. doi: 10.1161/JAHA.119.012447

12  Bowdish ME, D'Agostino RS, Thourani VH, *et al.* STS Adult Cardiac Surgery Database: 2021 Update on Outcomes, Quality, and Research. *Ann Thorac Surg*. 2021;111:1770–80. doi: 10.1016/j.athoracsur.2021.03.043

13  The Society of Thoracic Surgeons Expert Consensus for the Resuscitation of Patients Who Arrest After Cardiac Surgery. *Ann Thorac Surg*. 2017;103:1005–20. doi: 10.1016/j.athoracsur.2016.10.033

14  Nashef SAM, Roques F, Sharples LD, *et al.* EuroSCORE II†. *Eur J Cardiothorac Surg*. 2012;41:734–45. doi: 10.1093/ejcts/ezs043

15  Shahian DM, O'Brien SM, Filardo G, *et al.* The Society of Thoracic Surgeons 2008 cardiac surgery risk models: part 1--coronary artery bypass grafting surgery. *Ann Thorac Surg*. 2009;88:S2-22. doi: 10.1016/j.athoracsur.2009.05.053

16  O'Brien SM, Shahian DM, Filardo G, *et al.* The Society of Thoracic Surgeons 2008 cardiac surgery risk models: part 2--isolated valve surgery. *Ann Thorac Surg*. 2009;88:S23-42. doi: 10.1016/j.athoracsur.2009.05.056

17  Shahian DM, O'Brien SM, Filardo G, *et al.* The Society of Thoracic Surgeons 2008 cardiac surgery risk models: part 3--valve plus coronary artery bypass grafting surgery. *Ann Thorac Surg*. 2009;88:S43-62. doi: 10.1016/j.athoracsur.2009.05.055

18  Rankin JS, He X, O'Brien SM, *et al.* The Society of Thoracic Surgeons risk model for operative mortality after multiple valve surgery. *Ann Thorac Surg*. 2013;95:1484–90. doi: 10.1016/j.athoracsur.2012.11.077

19  Xue B, Li D, Lu C, *et al.* Use of Machine Learning to Develop and Evaluate Models Using Preoperative and Intraoperative Data to Identify Risks of Postoperative Complications. *JAMA Netw Open*. 2021;4:e212240. doi: 10.1001/jamanetworkopen.2021.2240

20  Wang C. Hierarchical Bayesian Modeling for Time-Dependent Inverse Uncertainty Quantification. 2024.

21  Li Y, Wang W, Yan X, *et al.* Research on the Application of Semantic Network in Disease Diagnosis Prompts Based on Medical Corpus. *Int J Innov Res Comput Sci Technol*. 2024;12:1–9. doi: 10.55524/ijircst.2024.12.2.1

22  Wang C, Wu X, Kozlowski T. Gaussian Process–Based Inverse Uncertainty Quantification for TRACE Physical Model Parameters Using Steady-State PSBT Benchmark. *Nucl Sci Eng*. 2019;193:100–14. doi: 10.1080/00295639.2018.1499279

23  Lee H-C, Yoon H-K, Nam K, *et al.* Derivation and Validation of Machine Learning Approaches to Predict Acute Kidney Injury after Cardiac Surgery. *J Clin Med*. 2018;7:322. doi: 10.3390/jcm7100322

24  Penny-Dimri JC, Bergmeir C, Reid CM, *et al.* Machine Learning Algorithms for Predicting and Risk Profiling of Cardiac Surgery-Associated Acute Kidney Injury. *Semin Thorac Cardiovasc Surg*. 2021;33:735–45. doi: 10.1053/j.semtcvs.2020.09.028

25  Li Y, Xu J, Wang Y, *et al.* A novel machine learning algorithm, Bayesian networks model, to predict the high-risk patients with cardiac surgery-associated acute kidney injury. *Clin Cardiol*. 2020;43:752–61. doi: 10.1002/clc.23377

26  Tseng P-Y, Chen Y-T, Wang C-H, *et al.* Prediction of the development of acute kidney injury following cardiac surgery by machine learning. *Crit Care*. 2020;24:478. doi: 10.1186/s13054-020-03179-9

27  Zea-Vera R, Ryan CT, Navarro SM, *et al.* Development of a Machine Learning Model to Predict Outcomes and Cost After Cardiac Surgery. *Ann Thorac Surg*. 2023;115:1533–42. doi: 10.1016/j.athoracsur.2022.06.055

28  Zhong Z, Yuan X, Liu S, *et al.* Machine learning prediction models for prognosis of critically ill patients after open-heart surgery. *Sci Rep*. 2021;11:3384. doi: 10.1038/s41598-021-83020-7

29  Xue B, Jiao Y, Kannampallil T, *et al.* Perioperative Predictions with Interpretable Latent Representation. *Proceedings of the 28th ACM SIGKDD Conference on Knowledge Discovery and Data Mining*. Washington DC USA: ACM 2022:4268–78.

30  Rajkomar A, Oren E, Chen K, *et al.* Scalable and accurate deep learning with electronic health records. *Npj Digit Med*. 2018;1:18. doi: 10.1038/s41746-018-0029-1

31  Miotto R, Li L, Kidd BA, *et al.* Deep Patient: An Unsupervised Representation to Predict the Future of Patients from the Electronic Health Records. *Sci Rep*. 2016;6:26094. doi: 10.1038/srep26094

32  Kingma DP, Welling M. Auto-encoding variational bayes. *ArXiv Prepr ArXiv13126114*. 2013.

33  Fritz BA, Chen Y, Murray-Torres TM, *et al.* Using machine learning techniques to develop forecasting algorithms for postoperative complications: protocol for a retrospective study. *BMJ Open*. 2018;8:e020124. doi: 10.1136/bmjopen-2017-020124

34  Rong X. word2vec parameter learning explained. *ArXiv Prepr ArXiv14112738*. 2014.

35  Bellini V, Valente M, Bertorelli G, *et al.* Machine learning in perioperative medicine: a systematic review. *J Anesth Analg Crit Care*. 2022;2:2. doi: 10.1186/s44158-022-00033-y

36  Chen T, Guestrin C. Xgboost: A scalable tree boosting system. *Proceedings of the 22nd acm sigkdd international conference on knowledge discovery and data mining*. 2016:785–94.

37  Pedregosa F, Varoquaux G, Gramfort A, *et al.* Scikit-learn: Machine learning in Python. *J Mach Learn Res*. 2011;12:2825–30.

38  Van Rossum G, Drake FL. *Python 3 Reference Manual*. Scotts Valley, CA: CreateSpace 2009.

39  Paszke A, Gross S, Massa F, *et al.* Pytorch: An imperative style, high-performance deep learning library. *Adv Neural Inf Process Syst*. 2019;32.

40  Chen RT, Li X, Grosse RB, *et al.* Isolating sources of disentanglement in variational autoencoders. *Adv Neural Inf Process Syst*. 2018;31.

41  Kim H, Mnih A. Disentangling by factorising. *International conference on machine learning*. PMLR 2018:2649–58.

42  Snell J, Swersky K, Zemel R. Prototypical networks for few-shot learning. *Adv Neural Inf Process Syst*. 2017;30.

43  Finn C, Abbeel P, Levine S. Model-agnostic meta-learning for fast adaptation of deep networks. *International conference on machine learning*. PMLR 2017:1126–35.

44  Zhao S, Song J, Ermon S. Infovae: Information maximizing variational autoencoders. *ArXiv Prepr ArXiv170602262*. 2017.

45  Sundararajan M, Taly A, Yan Q. Axiomatic Attribution for Deep Networks. In: Precup D, Teh YW, eds. *Proceedings of the 34th International Conference on Machine Learning*. PMLR 2017:3319–28.

46  Van der Maaten L, Hinton G. Visualizing data using t-SNE. *J Mach Learn Res*. 2008;9.

47  Brown TB. Language models are few-shot learners. *ArXiv Prepr ArXiv200514165*. 2020.

48  Lundberg SM, Lee S-I. A unified approach to interpreting model predictions. *Proceedings of the 31st International Conference on Neural Information Processing Systems*. Long Beach, California, USA: Curran Associates Inc. 2017:4768–77.

49  Abraham J, Bartek B, Meng A, *et al.* Integrating machine learning predictions for perioperative risk management: Towards an empirical design of a flexible-standardized risk assessment tool. *J Biomed Inform*. 2023;137:104270. doi: 10.1016/j.jbi.2022.104270

50  Abraham J, Meng A, Sona C, *et al.* An observational study of postoperative handoff standardization failures. *Int J Med Inf*. 2021;151:104458. doi: 10.1016/j.ijmedinf.2021.104458

51  Abraham J, King CR, Meng A. Ascertaining design requirements for postoperative care transition interventions. *Appl Clin Inform*. 2021;12:107–15.

52  Abraham J, Meng A, Tripathy S, *et al.* Systematic review and meta-analysis of interventions for operating room to intensive care unit handoffs. *BMJ Qual Saf*. 2021;30:513–24. doi: 10.1136/bmjqs-2020-012474